\documentclass[acmtog, margin=1in]{acmart}

\usepackage{booktabs} 

\usepackage{multirow}

\usepackage[ruled]{algorithm2e} 

\SetAlFnt{\small}
\SetAlCapFnt{\small}
\SetAlCapNameFnt{\small}
\SetAlCapHSkip{0pt}

\acmJournal{TOG}

\newcommand{\name}{Texture++}

\newcommand{\Skip}[1]{}

\usepackage{listings}
\lstdefinelanguage{json}{
    basicstyle=\ttfamily\small,
    numbers=left,
    numberstyle=\scriptsize\color{gray},
    stepnumber=1,
    numbersep=8pt,
    showstringspaces=false,
    breaklines=false,
    frame=none,
    morekeywords={type, path, pattern, show_as, group_by},
    keywordstyle=\bfseries\color{black},
    stringstyle=\color{black},
    columns=fullflexible
}

\begin{document}
\title{Texture++: Elevating 3D Asset Texture Resolution with a Region-Aware Diffusion Model}

\author{Shuaiwei Wang}
\affiliation{
  \institution{State Key Laboratory of CAD\&CG, Zhejiang University}
  \city{Hangzhou}
  \country{China}
}
\authornote{Equal Contribution.}

\author{Shi Li}
\affiliation{
  \institution{State Key Laboratory of CAD\&CG, Zhejiang University}
  \city{Hangzhou}
  \country{China}
}
\authornotemark[1]

\author{Jieting Xu}
\affiliation{
  \institution{State Key Laboratory of CAD\&CG, Zhejiang University}
  \city{Hangzhou}
  \country{China}
}

\author{Yuchi Huo}
\affiliation{
  \institution{State Key Laboratory of CAD\&CG, Zhejiang University}
  \city{Hangzhou}
  \country{China}
}

\author{Qi Wang}
\affiliation{
  \institution{North China Electric Power University}
  \city{Beijing}
  \country{China}
}

\author{Wenting Zheng}
\affiliation{
  \institution{State Key Laboratory of CAD\&CG, Zhejiang University}
  \city{Hangzhou}
  \country{China}
}

\author{Rengan Xie}
\affiliation{
  \institution{State Key Laboratory of CAD\&CG, Zhejiang University}
  \city{Hangzhou}
  \country{China}
}
\authornote{Corresponding Author.}



\begin{teaserfigure}
  \centering
	\includegraphics[width=1\textwidth]{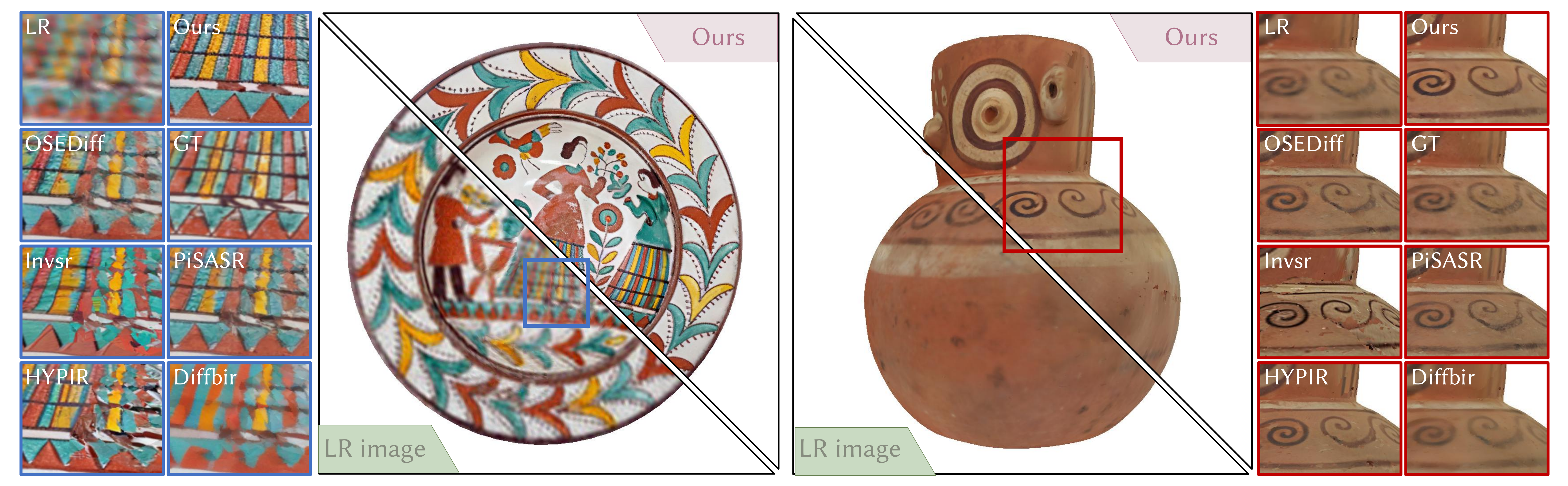}
    \caption{
    \name{} is capable of elevating the texture resolution of 3D assets. The visual comparison displays 3D meshes diagonally split between the low-resolution input (bottom-left) and our upscaled result (top-right). As highlighted by the zoomed-in regions, our approach recovers intricate details that are significantly closer to the ground truth, outperforming recent state-of-the-art methods. 
    } 
    \label{fig:teaser}
\end{teaserfigure}

\begin{abstract}

Numerous 3D assets are discarded due to low texture resolution, while current super-resolution models ignore texture maps and focus on natural images. An efficient and  generalizable texture super-resolution model can revitalize a large corpus of aging yet valuable assets across industries such as film and video games. We present \name, a novel framework for texture super-resolution, which enhances the low-resolution textures of assets to produce high-resolution, high-quality results. Specifically, we reformulate the task of super-resolution in UV space into performing it across multiple rendered views and merging the outputs. Firstly, to achieve more complete and continuous textures in the view space, we propose an adaptive view selection strategy to integrate textures dispersed across UV texture patches. Furthermore, we introduce a quadtree-based texture region organization method for combining super-resolved textures from different viewpoints, providing masks to distinguish regions that require improvement. Finally, we design a diffusion-based super-resolution model that enhances the texture resolution for specified masked regions, seamlessly integrating with surrounding regions. Through comprehensive evaluations, we demonstrate that our approach yields textures with substantially improved detail and coherence over existing methods.


\end{abstract}




\maketitle

\section{Introduction}
High-fidelity 3D assets are vital for modern digital experiences, underpinning immersive media, video games, virtual reality, and industrial digital twins.
Surface texture quality dictates the visual realism of 3D assets, yet crafting high-resolution (HR), artistically consistent textures is labor-intensive and costly. Many existing models have low-resolution (LR) textures, limiting their modern applicability. Efficient texture super-resolution (SR) techniques thus boost visual fidelity without asset re-creation, preserving artistic intent and compatibility with modern rendering pipelines.

\begin{figure}
    \centering
    \includegraphics[width=\linewidth]{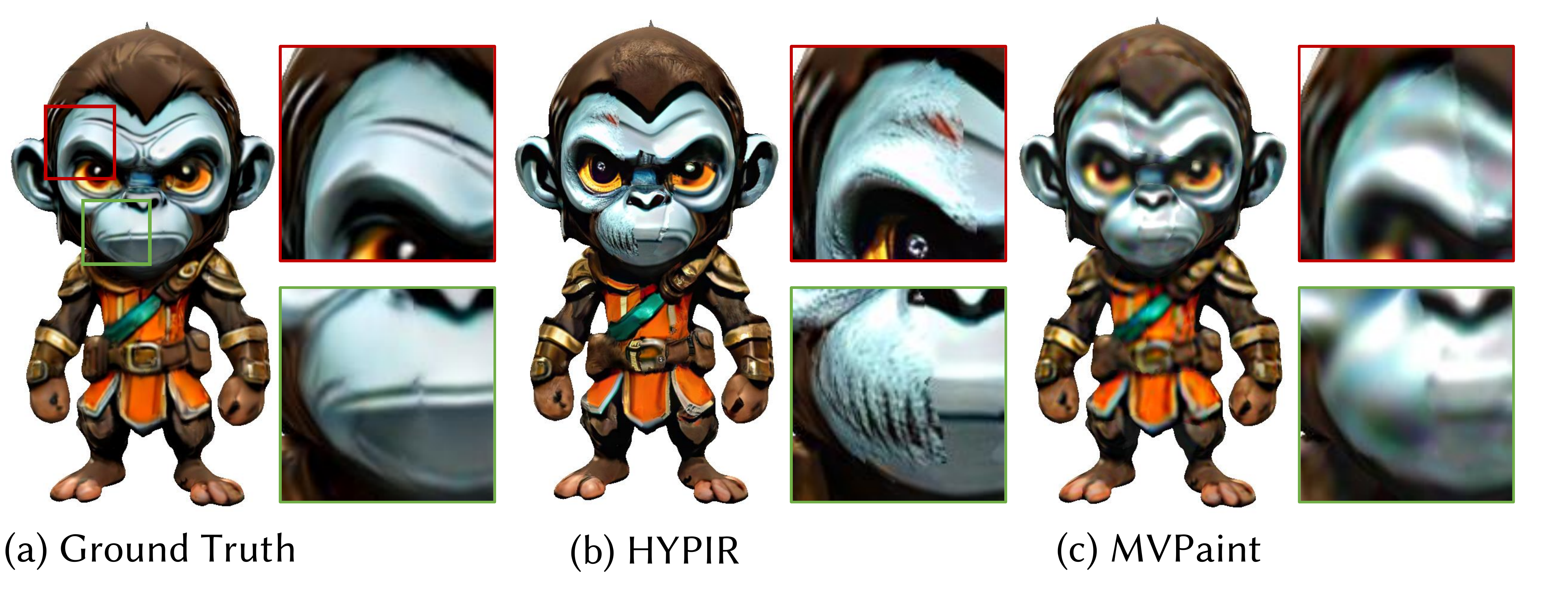}
    \vspace{-0.8cm}
    \caption{Failure cases of baseline methods on texture SR. (a) HR ground truth. (b) A naive image SR model applied directly to the texture produces severe seam artifacts across UV boundaries. (c) A texture generation method fails to preserve fine content and yields over-smoothed details.}
    \vspace{-0.3cm}
    \label{fig:problem}
\end{figure}

A straightforward idea is to apply the image SR models \cite{zhu2025oftsronestepflowimage, li2025diffusionsteprealworldsuperresolution, sun2024improvingstabilityefficiencydiffusion, sun2025pixellevelsemanticleveladjustablesuperresolution, wu2024seesrsemanticsawarerealworldimage, lin2024diffbirblindimagerestoration, wu2024onestepeffectivediffusionnetwork, yue2025arbitrarystepsimagesuperresolutiondiffusion, lin2025harnessingdiffusionyieldedscorepriors} directly to the texture map.
As shown in Figure \ref{fig:problem}, although these models demonstrate proficiency in restoring photorealistic details within natural images, they are inherently flawed when utilized for textures and fail to generate continuous patterns across the artificial boundaries of UV seams.
This failure arises from fundamental differences in the image distributions between texture maps and natural images. UV mapping breaks the continuity of patterns found in natural images, producing fragmented pattern pieces that mix with the texture background and result in a distinct image distribution. To mitigate the seams introduced by UV mapping, prior work~\cite{richard2020learnedmultiviewtexturesuperresolution, chen2025pbrsrmeshpbrtexture, ranade20223d, li20193d} employs a differentiable rendering pipeline to render view-specific observations of the texture map, producing seamless natural images in view space for SR. The SR rendered images is back-propagated through gradients to optimize the texture map. However, the same region of the texture map can be SR multiple times from different views, and inconsistencies among these results can blur that texture region, while gradient optimization incurs substantial overhead. Further, there exist end-to-end models~\cite{richard2020learnedmultiviewtexturesuperresolution, ranade20223d,li20193d} that operate texture maps directly without optimization, but these methods are trained only on specific texture map datasets and may lack the generalization ability to handle different kinds of UV unwrapping methods. 

Recently, mesh-based texture generation methods~\cite{ho2020denoisingdiffusionprobabilisticmodels, richardson2023texture,xiong2024texgaussian,10.1145/3731158} have attracted increasing interest and often adopt a generate-then-refine strategy. Their refinement stage mainly serves as a generation-side cleanup process, re-editing or inpainting synthesized textures to improve visual quality, prompt alignment, seam handling, and multi-view consistency. Unlike texture SR methods, it is not designed to recover high-resolution details from a given low-resolution texture, nor does it enforce strict content preservation or degradation consistency.


In this paper, we present \name, a novel texture SR method that is applicable to arbitrary UV mappings and enables the generation of high-resolution texture maps without any gradient-based optimization. Our core objective is to achieve the efficiency of end-to-end methods in texture space while preserving the strong generalization of view-space SR.
Initially, we propose an iterative framework that renders view-space images in multiple passes and performs texture detail enhancement.
Then, we introduce an adaptive view selection strategy. Since a complete pattern yields the best SR performance, we categorize viewpoints into two types, observation viewpoints and canonical viewpoints. The former aims to connect two UV charts that become contiguous in view space to produce a coherent pattern, while the latter aims to maximize the visibility of content that lies within a single UV chart in texture space.
Later, to prevent flickering caused by applying SR to the same texture region across multiple rendered views, we introduce an SR mask generation mechanism to identify, at each iteration, the regions in the view image that require SR, and a quadtree-based mechanism to manage masks with smooth boundaries across texture patches.
Finally, to update only local regions in view space with SR, we design a specialized local SR model that injects high-frequency details into the specified regions with a single diffusion step while blending seamlessly with the surrounding context. 
Our main contributions can be summarized as follows:
\begin{itemize}
\item We propose a novel texture SR method, \name, that produces high-quality SR texture maps for LR assets without gradient optimization. Experiments show our method significantly outperforms SOTA texure map SR methods in both quantitative and qualitative evaluations.
\item We introduce an adaptive view selection strategy that favors complete patterns for stronger SR performance while jointly accounting for texture patterns both within a single UV chart and across different UV charts.
\item We introduce an SR mask generation mechanism that identifies update regions across iterations and a quadtree-based scheme that ensures smooth mask boundaries across texture patches.
\item We design a specialized, local SR diffusion model performs targeted enhancement within a specified mask while preserving the surrounding content and seamless boundary transition.
\end{itemize}

\section{Related Work}
\label{sec:Related Work}
\paragraph{Single Image SR} Recent advances in single image SR are largely driven by generative models, which excel at synthesizing photorealistic details~\cite{ai2024dreamclearhighcapacityrealworldimage, chen2025adversarialdiffusioncompressionrealworld, qu2024xpsrcrossmodalpriorsdiffusionbased, wang2024exploitingdiffusionpriorrealworld, wang2023sinsrdiffusionbasedimagesuperresolution, zhu2025oftsronestepflowimage, li2025diffusionsteprealworldsuperresolution, sun2024improvingstabilityefficiencydiffusion, sun2025pixellevelsemanticleveladjustablesuperresolution, wu2024seesrsemanticsawarerealworldimage, lin2024diffbirblindimagerestoration, wu2024onestepeffectivediffusionnetwork, yue2025arbitrarystepsimagesuperresolutiondiffusion, lin2025harnessingdiffusionyieldedscorepriors}. 
A dominant research thrust focuses on improving efficiency and control, with methods like OFTSR~\cite{zhu2025oftsronestepflowimage}, FluxSR~\cite{li2025diffusionsteprealworldsuperresolution} and CCSR~\cite{sun2024improvingstabilityefficiencydiffusion} developing single-step inference pipelines, and PiSA-SR~\cite{sun2025pixellevelsemanticleveladjustablesuperresolution} enabling a tunable trade-off between fidelity and realism via a dual-LoRA~\cite{hu2021loralowrankadaptationlarge} structure.
Concurrently, enhancing semantic consistency remains a critical goal, addressed by methods such as SeeSR~\cite{wu2024seesrsemanticsawarerealworldimage} which leverages degradation-aware prompts. Another line of work targets arbitrary-scale upsampling, employing techniques that range from dynamic convolutions to novel representations like 2D Gaussian Splatting (GSASR)~\cite{chen2025generalized}.
State-of-the-art 2D image restoration methods, designed for holistic rectangular images, induce severe artifacts when applied directly to UV-partitioned textures.
In contrast, Texture++, purpose-built for the 3D domain, adopts canonical view selection and dynamic masking to achieve globally coherent, seamlessly integrated details across the mesh surface.
\paragraph{Mesh-based Texture Generation} Automatic 3D texturing has recently seen a surge in generative methods leveraging 2D diffusion models~\cite{ho2020denoisingdiffusionprobabilisticmodels}.
Approaches such as TEXTure~\cite{richardson2023texture} pioneered text-to-texture synthesis by iteratively painting a mesh from multiple camera views, while AlignTex~\cite{10.1145/3731158} focuses on transferring pixel-perfect details from multi-view images onto a mesh. 
Other works, like TexGaussian~\cite{xiong2024texgaussian}, have even proposed alternative 3D representations to ensure geometric consistency during PBR material generation from scratch. 
While these methods excel at scratch 3D asset generation, they fail to address the critical need to enhance the vast corpus of existing 3D models with legacy LR textures. 
This poses an orthogonal challenge, where the core objective is not synthesis but faithful preservation and enhancement of original artistic intent. 
Our work is specifically designed to address this challenge, framing high-fidelity texture SR as a task complementary to such generative methods.
\paragraph{Texture SR} 
Learning-based texture SR methods are divided into end-to-end and zero-shot models. 
For instance, Richard et al.~\cite{richard2020learnedmultiviewtexturesuperresolution} proposed an end-to-end model unifying multi-view redundancy and learned single-image priors in a single network for texture enhancement. 
In contrast, zero-shot models like PBR-SR~\cite{chen2025pbrsrmeshpbrtexture} leverage pre-trained 2D SR models as strong priors, iteratively refining textures by minimizing view-space losses via differentiable rendering and applying multi-view constraints to mitigate inconsistencies. 
However, this iterative optimization is prone to artifact accumulation.
Instead, we propose a canonical view selection algorithm for stable, conflict-free supervision and adopt a single-step masked diffusion model ensuring monotonically beneficial updates.
This design effectively mitigates the error accumulation prevalent in prior optimization-based methods.

\begin{figure*}
    \centering

    \includegraphics[width=\linewidth]{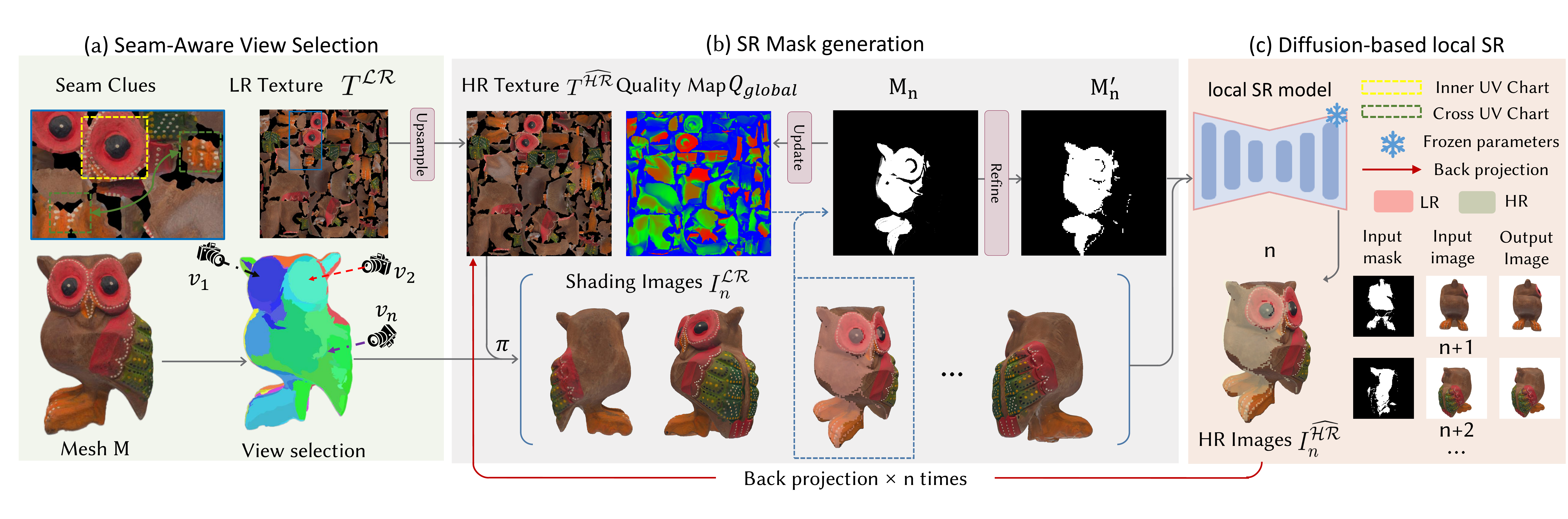}
    \vspace{-0.8 cm}
    \caption{\textbf{Overview.} Our method consists of four main stages. \textbf{(1) Observation view selection:} A set of canonical viewpoints $(v_1, \dots, v_n)$ are selected to render the 3D model $\mathcal{M}$. \textbf{(2) SR
region nomination:} For each view, we render a image $I^\mathcal{LR}_{n}$ and compute a per-view quality map $Q_n$. This is compared against a global quality map $Q_{global}$ to generate a binary mask $M_n$ that identifies regions for enhancement and updates the global quality map $Q_{global}$, which is then refined using quadtree-based regularization to produce the final mask $M_n'$. \textbf{(3) Local SR
generation:} The rendered image and its mask are inputed into the local SR model, which outputs a locally SR image $I^\mathcal{HR}_n$. \textbf{(4) Global texture update:} The new HR details are projected back onto the HR texture $T^\mathcal{HR}$.}
    \label{fig:overview} 
\end{figure*}

\section{Method}
\label{sec:Method}
\subsection{Overview}
Considering a single LR texture $T^\mathcal{LR}$, our goal is to recover the corresponding HR texture $T^\mathcal{HR}$.
Unlike traditional image SR, texture SR should incorporate the information of the 3D model $\mathcal{M}$ and UV mapping, otherwise it will yield poor recovery near the seam of UV charts.
For texture SR, with the camera $v_n$, the operator $\pi$ renders the geometric mesh with texture  $T$ on the image $I_{n} = {\pi}(\mathcal{M}, T, v_n)$.


To generate HR and view-consistent textures, we introduce a novel SR method through an iterative texture refinement framework. We initially use a simple bilinear upscaler to generate the HR texture $T^{\widehat{\mathcal{HR}}}$.
The pipeline of our method is illustrated in Figure~\ref{fig:overview} and can be divided into four modules:
1) \textbf{Observation view selection}. At first, we introduce a seam-aware observation view selection method to establish a set of canonical and consistent viewpoints.
2) \textbf{SR region nomination}. At each iteration, we take one view $v_n$ from the selected views and render the image $I_{n}$ with the HR texture $T^{\widehat{\mathcal{HR}}}$. Meanwhile, we generate a SR mask $M_n$ with $I_{n}$ and a global quality map, $Q_{global}$, which indicates texture regions that require resolution enhancement.
3) \textbf{Local SR generation}. Then we generate the locally SR image $\tilde{I}_n$ using a pretrained local SR model, conditioned on the rendered image $I_{n}$ and the mask $M_n$.
4) \textbf{Global texture update}. Finally, we seamlessly project the generated details $\tilde{I}_n$ back and update the predicted HR texture $T^{\widehat{\mathcal{HR}}}$.


\begin{figure}
    \centering
    \includegraphics[width=\linewidth]{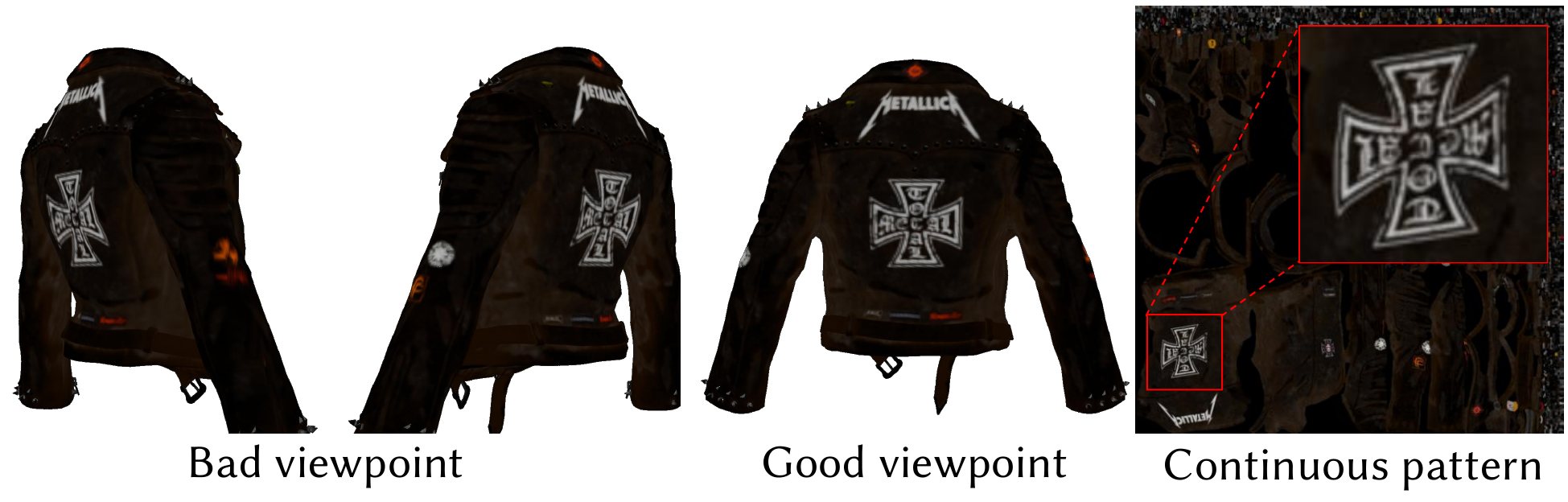}
    \vspace{-.6cm}
    \caption{\textbf{Effect of viewpoint quality on texture continuity.} Bad viewpoints (left) force a continuous surface pattern (right) to be updated across multiple views, disrupting its original continuity. In contrast, a good viewpoint (middle) covers the entire pattern, enabling it to be super-resolved consistently in a single view.}
    \label{fig:view selection}
    \vspace{-0.5 cm}

\end{figure}

\subsection{Seam-Aware View Observation Selection}
A significant challenge in texture SR is handling the artificial boundaries introduced by UV unwrapping. 
UV unwrapping cuts a continuous 3D surface into multiple isolated charts so that surface-adjacent regions may be far apart in the 2D texture map, as shown in Figure~\ref{fig:overview}.
This separation prevents standard image-based SR operations from naturally preserving cross-seam consistency. 
In addition, multi-view texture updates may assign different high-frequency details to the same surface area because of viewpoint-dependent visibility and projection differences, as shown in Figure~\ref{fig:view selection}.
Such inconsistent updates often result in spurious high-frequency artifacts and visible seams within UV charts and across their boundaries.
To address this issue, we introduce an adaptive observation view selection strategy.
The core idea is to assign a set of viewpoints to richly textured region to ensure stable and coherent convergence during iterative refinement.

Our strategy decouples viewpoint selection into two complementary schemes: one that preserves intra-chart coherence and another that enforces cross-seam texture continuity.
\paragraph{Viewpoint selection within the UV chart.}
We segment the mesh faces into distinct sets, where each set corresponds to a contiguous UV chart.
For each UV chart ${P}$, the camera's viewing direction is set to align with the area-weighted average normal of the corresponding mesh surface, ensuring a frontal perspective that maximizes the sampling density and minimizes projective distortion.
The camera is centered on the geometric centroid of the corresponding mesh surface.
The canonical view vector $v_P^*$ is computed as:
\begin{equation}
\label{eq:avg_normal}
v_P^* = \frac{\sum_{f_i \in P} A_i \cdot N_i}{\left| \sum_{f_i \in P} A_i \cdot N_i \right|_2}
\end{equation}
where $f_i$ is a face within the chart $P$, $A_i$ is area, and $N_i$ is normal. 
By precomputing a canonical viewpoint for each chart, we ensure that every UV chart’s texture pattern is covered by at least one viewpoint and that this viewpoint observes the chart as completely as possible, minimizing unobserved regions.
\paragraph{Viewpoint selection across the UV chart.} 
We define a seam as a geometric edge on the 3D mesh whose constituent vertices map to different pairs of vertices in the UV coordinate space. 
After identifying the set of all seam edges, we group them into contiguous seam $strips$, and each strip represents the continuous seam edges on the 3D model.
For each identified seam strip, we define an encompassing seam region, $\mathcal{R}_{\textit{s}}$, which is comprised of all mesh faces adjacent to any edge within the strip. 
Similarly to the intra-chart case, the camera's viewing direction is aligned with the area-weighted average normal of all faces in $\mathcal{R}_{\textit{s}}$. 
The camera is then positioned to be centered over the geometric centroid of this region.
\paragraph{Viewpoint Consolidation.}
To mitigate the computational overhead caused by numerous UV islands and seams, we use a hierarchical viewpoint consolidation strategy. 
We first apply area-weighted $k$-means clustering in normal space to group UV islands and seam strips into compact representative viewpoints, respectively. 
These candidates are then jointly merged via hierarchical clustering~\footnote{Please refer to the supplementary document for the details.}.
Finally, we reapply visibility culling to recompute the exact sets of visible faces.
This step effectively prunes inter-view redundancies while guaranteeing complete surface coverage.
Conversely, for models with simplistic UV layouts, such as a sphere that unwraps to a single large chart, a single canonical viewpoint would provide insufficient coverage for iterative refinement. 
In such cases, we render six axis-aligned views (front, back, top, bottom, left, right) of the model's bounding box.

\subsection{SR Region Nomination}
\label{subsec:mask generation}
Following the determination of canonical viewpoints, the next stage involves rendering the corresponding regions and identifying precisely which pixels within these renderings are suitable for SR and subsequent texture updates. 
A naive strategy of updating all rendered pixels is problematic for two reasons. First, diffusion models inherently introduce stochastic variations, repeatedly updating the same texel with slightly different outputs gradually blurs the texture. Second, indiscriminately re‑generating regions that are already of sufficient quality wastes computation and slows down the pipeline. 
To prevent those, we introduce an SR mask generation mechanism that identifies update regions across iterations, ensuring that texture updates improve monotonically. 
This is orchestrated through a global quality map that tracks the best-achieved rendering quality for every texel across all iterations.
\paragraph{Quality Evaluation.}
For each viewpoint $v_n$, we compute a corresponding geometric quality map $Q_{n}$.
The quality score at each pixel is determined by both the geometric orientation and the proximity of the camera to the surface.
The goal is to favor camera poses that capture the surface from a frontal perspective at close range, as these views maximize visible detail and minimize projective distortion.
We quantify this by weighting surface patches viewed frontally more heavily, where texture is sampled without foreshortening, and by penalizing distant views, where captured detail falls off inversely with the square of distance.
Concretely, for each pixel $(u, v)$, the value is:
\begin{equation}
    Q_{n}(u,v) = \frac{\max(0, \cos{(v^*\cdot N))}}{d^2}
\end{equation}
where $d$ is the distance from the camera to the surface, $v^*$ is the viewing direction and $N$ is the surface normal.
This formulation ensures that a high quality score is assigned only when the surface is both facing the camera directly and close to the camera. This dual criterion prevents distant views, which may suffer from low sampling density, from overwriting details captured by closer, higher-fidelity observations.
\paragraph{Mask Generation.}
To ensure that the iterative refinement process is strictly progressive, we maintain a global quality map, $Q_{global}$, which has the same dimensions as the HR texture. $Q_{global}$ stores the maximum rendering quality score achieved from any viewpoint across all preceding iterations for each texel.
We define a binary update mask in the image space, $M$, that indicates corresponding pixels require SR.
For each view, we first compute a quality map $Q_n(x,y)$ for the current rendered image, then retrieve the corresponding historical best scores $Q_{hist}(x, y)$ from the $Q_{\textit{global}}$. We then generate a raw mask, $M_{raw}$ by comparing the $Q_{view}(x,y)$ and $Q_{hist}(x,y)$:
    \begin{equation}
    M_{raw}(x,y) = \begin{cases}
        1 & Q_{\textit{hist}}(x,y) \le Q_{\textit{n}}(x,y) \\
        0 & Q_{\textit{hist}}(x,y) > Q_{\textit{n}}(x,y).
    \end{cases}
\end{equation}

Raw masks $M_{raw}$ are often produced with fragmented, jagged and irregular boundaries, which tend to be misinterpreted by the diffusion model as texture patterns, causing boundary artifacts. To address this, we introduce a quadtree-based mask regularization strategy that produces clean, structured mask boundaries.
Specifically, we construct a quadtree on the spatial domain of the rendered image, where the root represents the entire image. 
We then recursively subdivide each node into four equal-sized children and terminate under one of the following conditions: 
a) All pixels within the node's corresponding region in $M_{raw}$ are 0 (no SR required).
b) All pixels within the node's region in $M_{raw}$ are 1 (full SR required). 
c) The node has reached a predefined minimum size, signaling a boundary region where 0 and 1 pixels are mixed; we set all its pixels to 1 to consolidate these fragments into a clean block.
The mask $M_n$ provides a coherent and more stable signal to the diffusion model, facilitating the synthesis of high-quality details.
\begin{figure}
    \centering
    \includegraphics[width=\linewidth]{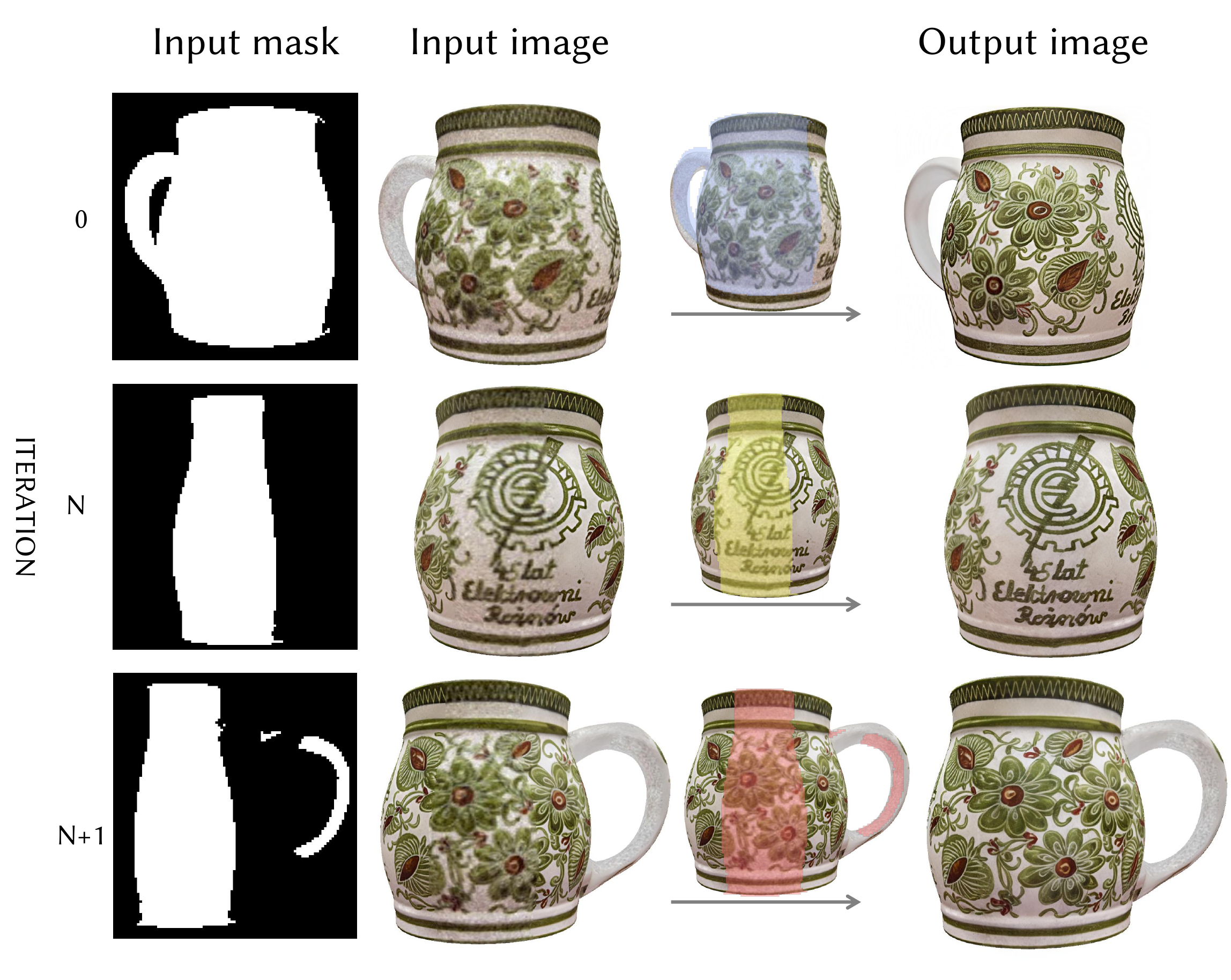}
    \vspace{-1 cm}
    \caption{\textbf{Visualization of the iterative refinement process.} The rows show snapshots of the pipeline at iterations 0, $N$ and $N + 1$. For each step, we present the rendered view of the current texture(\textbf{middle}), the mask identifying regions for update(\textbf{left}), and the locally SR output(\textbf{right}).}
    \label{fig:refinement process}
    \vspace{-0.6 cm}
\end{figure}

\subsection{Local SR Generation}
After identifying the local regions that require SR, the next step is to select an appropriate SR model. 
Our input at this stage is a rendered image that is sharp everywhere except in the target regions, which remain blurry. 
Image SR models are not designed for such a setting: they process the entire image uniformly, altering the sharp context and failing to harmonize the enhanced details with the surrounding texture, leaving visible seams or inconsistencies as shown in Figure~\ref{fig:ablation} (left). To address this, we fine-tune a pre-trained Stable Diffusion 2.1-base model (SD)~\cite{rombach2022highresolutionimagesynthesislatent} to SR local regions defined by our generated masks, ensuring seamless integration with the surrounding context.

First, to enable precise control over the generation process, we adapt its architecture to accept both an image and the corresponding mask $M$ as conditional inputs. 
The rendered image is encoded into a latent representation $z \in \mathbb{R}^{4 \times H \times W}$ using a VAE encoder. 
To restrict the SR to the masked regions, we provide the mask as an explicit spatial condition to the diffusion model. 
Concretely, we concatenate the latent $z$ with the mask $M$ along the channel dimension, creating a 5-channel input, $[z; M] \in \mathbb{R}^{5 \times H \times W}$, and modify the first convolutional layer of the U-Net accordingly.
Second, we leverage Low-Rank Adaptation (LoRA)~\cite{hu2021loralowrankadaptationlarge}. 
We integrate two trainable LoRA modules into the U-Net, one to restore high-fidelity content and the other to focus on specific areas. 
Inspired by some previous advancements in SR~\cite{sun2025pixellevelsemanticleveladjustablesuperresolution, 7839189, zhang2018residualdensenetworkimage}, we frame the generation as a residual prediction problem in the latent space.
Specifically, the U-Net is trained to predict the latent space residual, $\Delta z$, required to transform the initial latent $z$ into the high-fidelity latent $z_{final}$, which is computed as: 
\begin{equation}
    z_{final} = z - \Delta z \odot M
\end{equation}
where $\odot$ denotes the Hadamard (element-wise) product, $M$ is the binary mask. This residual formulation provides a stable training target, allowing for a direct, single-step SR without the need for an iterative denoising schedule.
\paragraph{Model Optimization.} 
To ensure that the generated details are not only structurally accurate but also perceptually convincing, we train our model using a hybrid loss function. 
The loss $\mathcal{L}$ is a weighted combination of a masked pixel-wise Mean Squared Error (MSE) loss $\mathcal{L}_{mse}$ and the masked Learned Perceptual Image Patch Similarity (LPIPS)~\cite{zhang2018unreasonableeffectivenessdeepfeatures} loss $\mathcal{L}_{lpips}$  in the decoded image space.
\begin{equation}    
    \mathcal{L}_{mse} = \lVert (\mathcal{D} (z_{final}) - \mathbf{I}_{GT}) \odot M \rVert ^2
\end{equation}
where $\mathcal{D}(\cdot)$ is the VAE decoder, $\mathbf{I}_{GT}$ is the GT image.
Similarly, the masked perceptual loss, $\mathcal{L}_{lpips}$, leverages a pre-trained network VGG~\cite{simonyan2015deepconvolutionalnetworkslargescale}to compare features in a perceptually relevant space, again focusing on the masked region:
\begin{equation}
    \mathcal{L}_{lpips} = LPIPS(\mathcal{D}(z_{final}) \odot M, \mathbf{I}_{GT} \odot M)
\end{equation}
The final training objective is the weighted sum of these two losses:
\begin{equation}
    \mathcal{L} = \lambda_{mse} \mathcal{L}_{mse} + \lambda_{lpips} \mathcal{L}_{lpips}
\end{equation}
\vspace{-0.5cm}
\paragraph{Inference for Texture Refinement.} 
During the inference phase of our pipeline, the rendered image $I$ from a viewpoint and its corresponding mask $M$ serve as inputs to our local SR model,
which enables the SR module to utilize the high-frequency context from previous iterations, ensuring global consistency.
The output is a locally SR image that exhibits rich detail in the target region and maintains a seamless boundary with its surroundings.
\subsection{Global Texture Update}
The final stage of each iteration is to project the locally SR image, generated by our model, back onto the HR texture $T^{\widehat{\mathcal{HR}}}$.
Our framework accomplishes this through an direct projection mechanism~\cite{zeng2024paint3d}, avoiding slower optimization-based techniques~\cite{youwang2024paint, chen2025pbrsrmeshpbrtexture}.
The update is constrained by the binary mask $M_n$ (Sec \ref{subsec:mask generation}), which ensures that only regions with improved visual quality are modified. Figure \ref{fig:refinement process} visualizes this process, illustrating how our method dynamically targets and refines different surface regions across iterations to cumulatively build a high-fidelity 3D asset.

\begin{figure*}
    \centering
    \includegraphics[width=\linewidth]{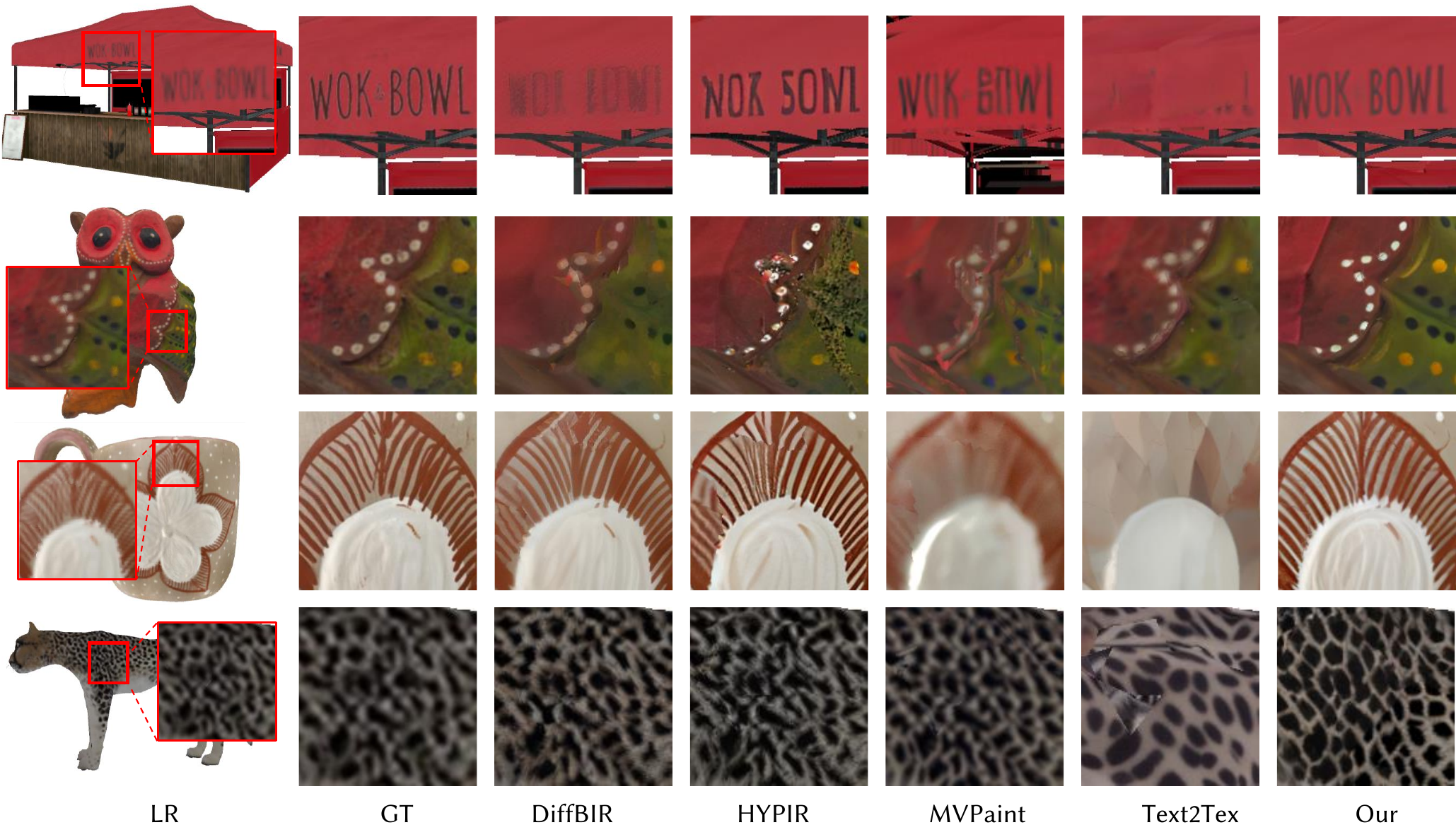}
    \caption{Qualitative comparison with SOTA image SR (DiffBIR, HYPIR) and texture generation methods (MVPaint, Text2Tex). The image SR baselines produce jarring seam artifacts across UV boundaries, while the texture generation methods tend to over-smooth details and erase fine text and structural patterns. In contrast, our method reconstructs crisper textures and avoids these boundary discontinuities.}
    \label{fig:baseline}
    \vspace{-0.2cm}
\end{figure*}
\section{Experiments}
In this section, to validate the effectiveness of our method, we conduct a series of comprehensive experiments. We first detail the implementation and training setup. Then, we present both qualitative and quantitative comparisons against several state-of-the-art (SOTA) methods. Finally, we perform ablation studies to analyze the impact of key components in our framework.
\subsection{Implementation Details}
\label{subsec:implementation details}
Our entire framework is implemented in PyTorch. For the core task of rendering, we leverage the Nvidiffrast~\cite{laine2020modularprimitiveshighperformancedifferentiable}. We use three NVIDIA A6000 GPUs with 48 GB of VRAM for SR model training and one NVIDIA A6000 GPU for iterative texture refinement. We train our model on a custom pairwise local SR dataset, and more details on the dataset and training procedure can be found in the supplementary material.
\begin{table}[htbp]
    \centering
      \begin{tabular}{l c c c c c}
        \toprule
        Method & PSNR $\uparrow$ & SSIM $\uparrow$ & LPIPS $\downarrow$ & DISTS $\downarrow$ & Time(s) $\downarrow$\\
        \midrule
        DiffBIR & \textcolor{blue}{35.9406} & \textcolor{blue}{0.9292} & 0.1067 & 0.1056 & 586.2\\
        HYPIR & \textcolor{green}{33.3778} & 0.8997 & 0.1066 & 0.0996 & \textcolor{blue}{48.0}\\
        OSEDiff & 31.6975 & \textcolor{green}{0.9168} & \textcolor{green}{0.0959} & \textcolor{green}{0.0964} & 108.6\\
        PiSASR & 33.1539 & 0.9159 & \textcolor{blue}{0.0866} & \textcolor{blue}{0.0864} & 111.4 \\
        InvSR & 30.5382 & 0.8789 & 0.1281 & 0.1123 & \textbf{\textcolor{red}{42.0}}\\
        Text2Tex & 27.5001 & 0.8855 & 0.1405 & 0.1272 & 321.5\\
        Paint3D & 24.3583 & 0.8300 & 0.1826 & 0.1570 & 130.4\\
        MVPaint & 20.9008 & 0.7098 & 0.2581 & 0.1872 & 215.6\\
        \textbf{Our} & \textbf{\textcolor{red}{37.5277}} & \textbf{\textcolor{red}{0.9524}} & \textbf{\textcolor{red}{0.0637}} & \textbf{\textcolor{red}{0.0736}} & \textcolor{green}{94.4}\\
        \bottomrule
    \end{tabular}
    \caption{Quantitative comparison with state-of-the-art methods for 4x texture SR. We report PSNR, SSIM, LPIPS, and DISTS. Best results are in \textcolor{red}{red}, second best in \textcolor{blue}{blue}, and third best in \textcolor{green}{green}.}
    \label{tab:quantitative_results}
    \vspace{-0.8cm}
\end{table}
\subsection{Comparison}
Due to the scarcity of work related to texture SR and the existing methods are not open-source~\cite{chen2025pbrsrmeshpbrtexture, richard2020learnedmultiviewtexturesuperresolution}, we conduct extensive comparisons against a suite of SOTA image SR and texture generation methods.
We select five powerful and contemporary image SR baseline models, representing the cutting edge of diffusion-based and efficient SR. HYPIR\cite{lin2025harnessingdiffusionyieldedscorepriors}, a method that harnesses diffusion-yielded score priors for high-quality image restoration. DiffBIR\cite{lin2024diffbirblindimagerestoration}, another strong diffusion-based model that uses a dual-branch architecture for controlled image restoration. OSEDiff \cite{wu2024onestepeffectivediffusionnetwork} uses one-step effective diffusion network for faster inference. InvSR\cite{yue2025arbitrarystepsimagesuperresolutiondiffusion}, an efficient framework for arbitrary-step SR based on invertible neural networks. PiSASR\cite{sun2025pixellevelsemanticleveladjustablesuperresolution}, an advanced single-step SR model that also utilizes a pre-trained Stable Diffusion backbone.
These methods are designed for 2D image SR and cannot be directly integrated as a local SR module in our pipeline, since our iterative refinement process generates hybrid-resolution rendering images that comprise both sharp and blurry regions, which is incompatible with standard SR methods designed for uniformly LR inputs. Therefore, we apply each baseline to the entire LR texture to generate a HR output in a single pass.

In addition, we select three texture generation methods that employ coarse-to-fine strategies with dedicated refinement stages. Text2Tex~\cite{chen2023text2tex} progressively synthesizes HR partial textures from multiple viewpoints using a depth-aware inpainting diffusion model. Paint3D~\cite{zeng2024paint3d} first generates coarse texture maps through multi-view texture fusion, then refines them using UV Inpainting and UVHD diffusion models in UV space. MVPaint~\cite{cheng2025mvpaint} introduces a three-stage framework with synchronized multi-view generation, spatial-aware 3D inpainting, and UV refinement that performs SR in UV space followed by spatial-aware seam smoothing.
Since these methods are designed for texture generation from scratch rather than SR of existing textures, we adapt them to our task by utilizing their refinement stages. Specifically, we use the LR textures as input to their refinement modules and perform the same refinement operations as in their original pipelines.

For all baselines, we use their officially released pre-trained models and follow the recommended inference settings to ensure optimal performance. All experiments are conducted for a 4$\times$ upsampling factor, partial results are shown in Figure \ref{fig:baseline} and Table \ref{tab:quantitative_results}, with other results available in Figure \ref{fig:another baseline} in the supplementary material.
\paragraph{Qualitative Comparison.}
As shown in Figure \ref{fig:baseline}, ~\ref{fig:baseline2} and Figure \ref{fig:another baseline}, while image SR baselines may produce sharper details, their lack of 3D spatial awareness leads to severe seams and boundary artifacts. Texture generation methods fail to preserve the original content or achieve SR, resulting in blurred, distorted, or even erased text and structural details.
On the beige mug (row 3) and wooden owl (row 2), baselines introduce spurious noise and distort original patterns. Text on the scene model (row 1) is either blurred, corrupted by artifacts in all baseline methods, whereas ours reconstructs crisp, readable characters. In contrast, our 3D-aware approach reliably synthesizes high-fidelity details and ensures seamless consistency across all views, yielding results closest to the GT.
\paragraph{Quantitative Comparison.} We provide a quantitative evaluation using four standard image quality assessment metrics: Peak Signal-to-Noise Ratio (PSNR), Structural Similarity Index (SSIM)\cite{1284395}, LPIPS\cite{zhang2018unreasonableeffectivenessdeepfeatures}, and DISTS\cite{Ding_2020}. The metrics are computed between the Output HR texture and the GT texture. As presented in Table \ref{tab:quantitative_results}, our method achieves the best performance in the all metrics. This indicates a superior ability to accurately reconstruct the GT structure and content, which aligns with our qualitative findings that baseline methods often hallucinate incorrect details. 
We also report time cost. Although our pipeline involves an iterative process, it still takes less time than most methods. This is possibly due to the fact that  we SR rendered images in view space (1K), rather than directly operating on the texture (2K). 
More importantly, our SR model is a single-step diffusion model that performs SR on rendered images efficiently. In contrast, baseline methods typically process texture at the target resolution (2K).
\begin{table}[htbp]
    \centering
    \begin{tabular*}{\columnwidth}{l @{\extracolsep{\fill}} cccc}
        \toprule
        Setting & PSNR $\uparrow$ & SSIM $\uparrow$ & LPIPS $\downarrow$ & DISTS $\downarrow$ \\
        \midrule
        w/o v.s. & 36.3777 & 0.9545 & 0.0482 & 0.0875 \\
        w/o quadtree & 38.3475 & 0.9685 & 0.0435 & 0.0765 \\
        w/o mask & 36.2000 & 0.9641 & 0.0490 & 0.0979 \\
        \midrule
        \textbf{full} & \textbf{38.4241} & \textbf{0.9686} & \textbf{0.0433} & \textbf{0.0760} \\
        \bottomrule
    \end{tabular*}
    \caption{\textbf{Quantitative results of the ablation study.} We report metrics for variants of our method, including removing our view selection (v.s.), quadtree processing, and masked synthesis.}
    \label{tab:ablation_results}
    \vspace{-1cm}
\end{table}
\subsection{Ablation Study}
\label{sec:Ablation Study}
\begin{figure*}
    \centering
    \includegraphics[width=\linewidth]{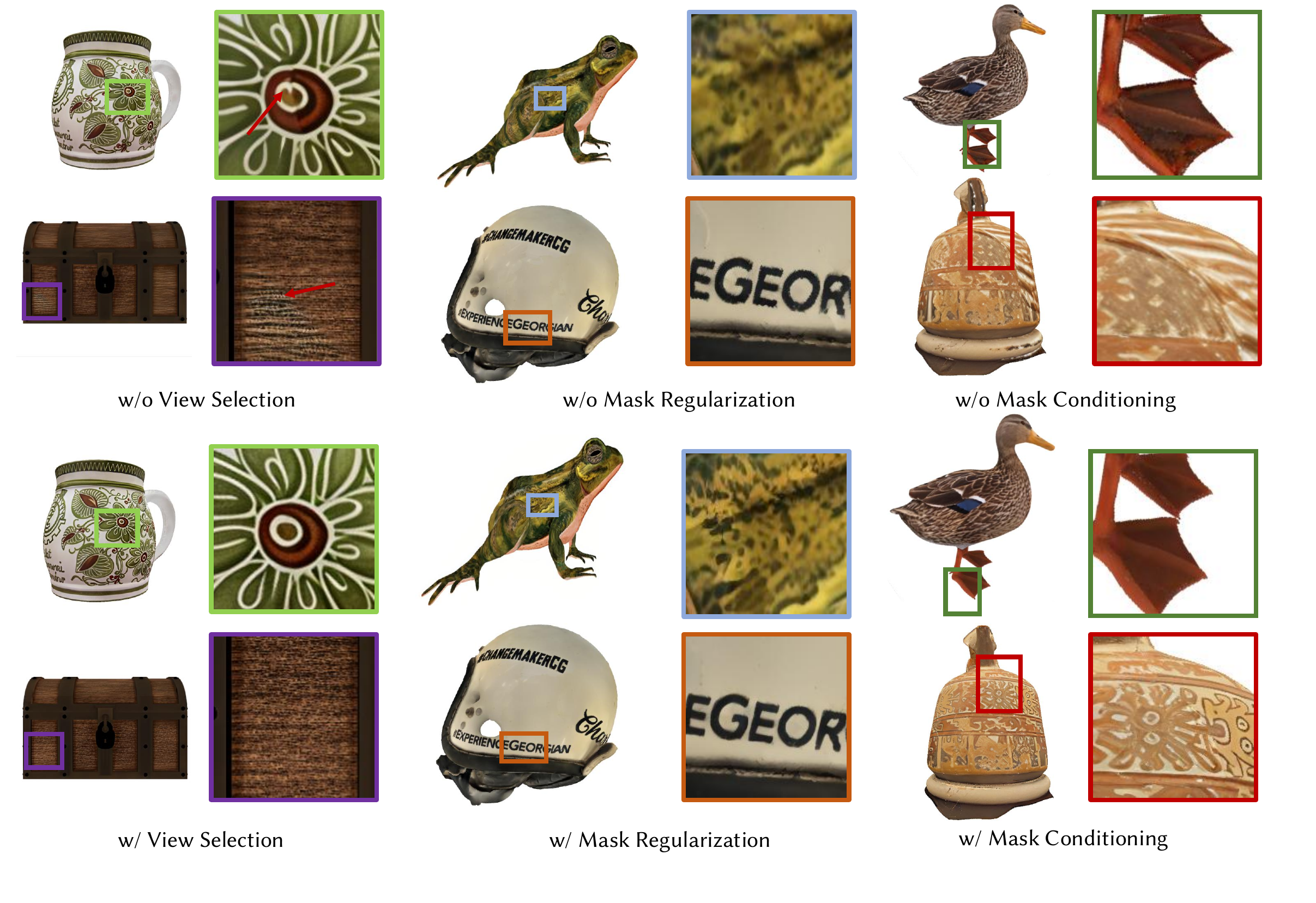}
    \vspace{-1.5cm}
    \caption{\textbf{Ablation study of the key components.}We visualize the impact of each proposed module by comparing the ablated baselines (top) with our full method (bottom). The results demonstrate that consistent view selection is crucial for geometric continuity, mask regularization is essential for detail sharpness, and mask conditioning is indispensable for seamless integration.}
    \label{fig:ablation}
    
\end{figure*}

We conduct ablation studies to evaluate the key components of our framework. Specifically, we analyze the impact of our observation view selection, quadtree-based mask processing, and local SR model.
The results, presented in Table \ref{tab:ablation_results} and Figure \ref{fig:ablation}, demonstrate the necessity of each component for achieving high-quality results.
\paragraph{Effect of Observation View Selection.} To validate the view selection strategy, we compare it against a baseline that employs random viewpoint sampling. As illustrated in the Figure \ref{fig:ablation} (left), this naive approach fails to preserve topological continuity. Without canonical views optimized to frame UV boundaries, the model generates inconsistent updates for adjacent texture regions, resulting in severe seam artifacts that sever continuous patterns across the 3D surface. In contrast, our strategy ensures that updates are geometrically aligned, producing globally coherent textures.
\paragraph{Effect of Quadtree-based Mask Processing.} Raw, pixel-level comparisons often yield fragmented masks with high-frequency noise along the boundaries. To verify the necessity of our quadtree regularization, we ablate this component and use the raw masks directly. As shown in Figure \ref{fig:ablation} (middle), this leads to a noticeable degradation in sharpness and detail. The noisy, disconnected mask regions fail to provide a coherent spatial guide, preventing the SR model from effectively focusing its generative capacity. By regularizing these regions via quadtree decomposition, our full method provides a robust spatial prior, enabling the synthesis of significantly sharper and more intricate textures.

\paragraph{Effect of Local SR} Finally, we validate the importance of our local SR model, which takes the mask as an explicit spatial condition. We compare our approach against a baseline where the SR module is replaced by a standard SR model~\cite{sun2025pixellevelsemanticleveladjustablesuperresolution}. In this setup, masking is applied after the entire image has been SR. As shown in  Figure \ref{fig:ablation} (right), this strategy invariably produces pronounced boundary artifacts at the interface with the existing texture. Lacking mask-based conditioning during the generative process, the baseline model is unconstrained by the surrounding context, resulting in severe color and pattern discontinuities at the patch periphery. In contrast, our model explicitly conditions the synthesis on the mask, ensuring that the generated content is contextually coherent and seamlessly integrated.

\section{Conclusion}

In conclusion, we presented \name, a texture SR method that supports arbitrary UV mappings and produces HR textures. We introduced an iterative framework that refines texture detail via multi-view rendering, with an adaptive view selection strategy that uses observation views to connect adjacent UV charts for coherent patterns and canonical views to maximize visibility within a single chart. We further developed an SR mask generation scheme with a quadtree-based representation that selects regions requiring SR at each iteration and preserves smooth boundaries across texture patches, and we designed a local SR diffusion model that updates only targeted regions in view space, injecting high-frequency details while blending seamlessly with surrounding context.
Experiments validate our approach, as quantitative metrics show significant improvements over SOTA baselines and qualitative results confirm that the framework can generate seamless and HR textures. We discuss the limitations and potential directions for future work in the Supplementary Material.

\bibliographystyle{ACM-Reference-Format}
\bibliography{main}




\begin{figure*}
    \centering
    \includegraphics[width=\linewidth]{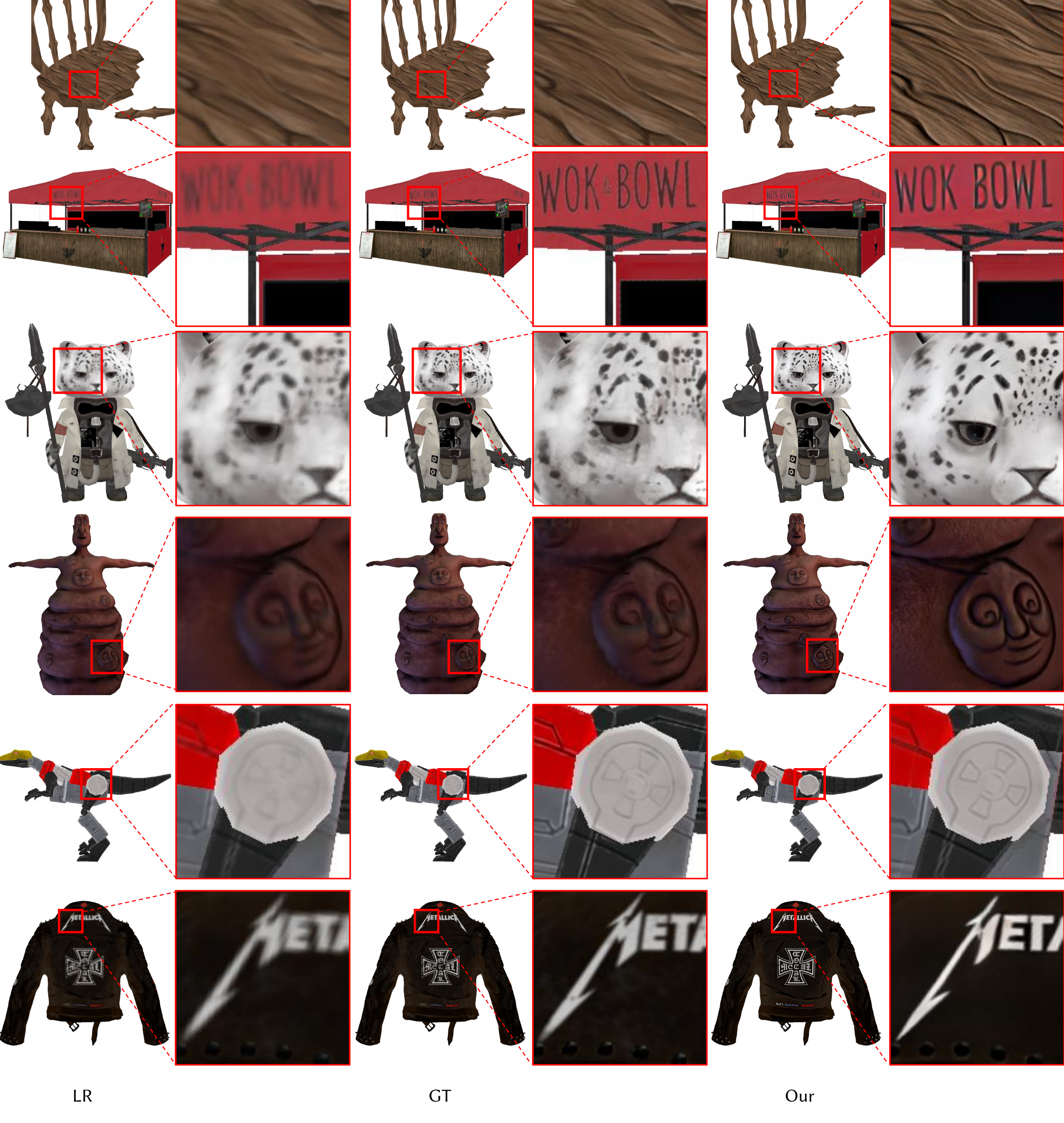}
    \caption{\textbf{Qualitative results (1/3).} Each triplet shows the rendered LR input, our HR output, and the GT. In these examples, our method excels at restoring sharp text and fine textural details that are often severely degraded in the LR inputs.}
    \label{fig:more results1}
\end{figure*}

\begin{figure*}
    \centering
    \includegraphics[width=\linewidth]{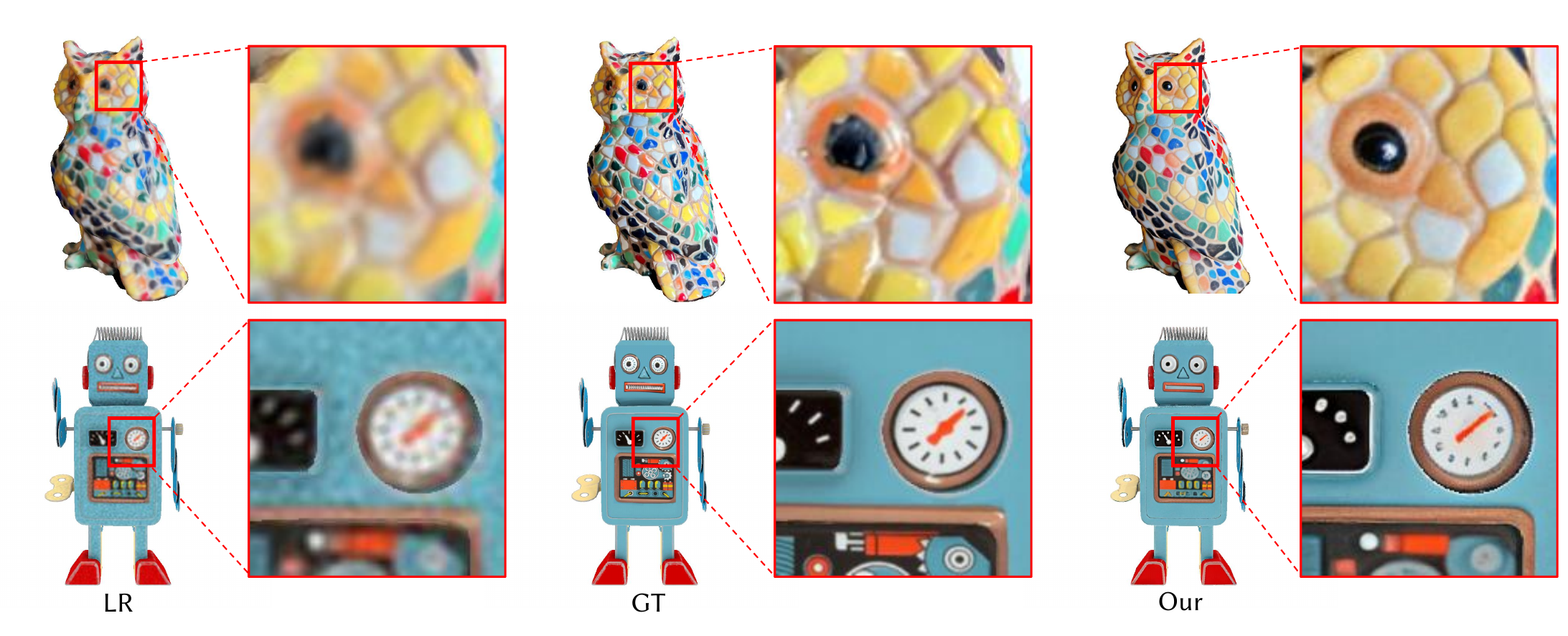}
    \vspace{-1.0cm}
    \caption{\textbf{Qualitative results (2/3).} Each triplet shows the rendered LR input, our HR output, and the GT. These cases further demonstrate our method's ability to recover fine surface details and maintain texture consistency.}
    \label{fig:more results2}
\end{figure*}

\clearpage
\appendix

\twocolumn[
\begin{center}
    {\huge \bf Texture++: Elevating 3D Asset Texture Resolution with a Region-Aware Diffusion Model}\\
     \vspace{0.3cm}
    {\LARGE \emph{Supplementary Material}}
     \vspace{0.4cm}
\end{center}
]

\section{Limitations and Future Work.} 
\begin{figure}
    \centering
    \includegraphics[width=\linewidth]{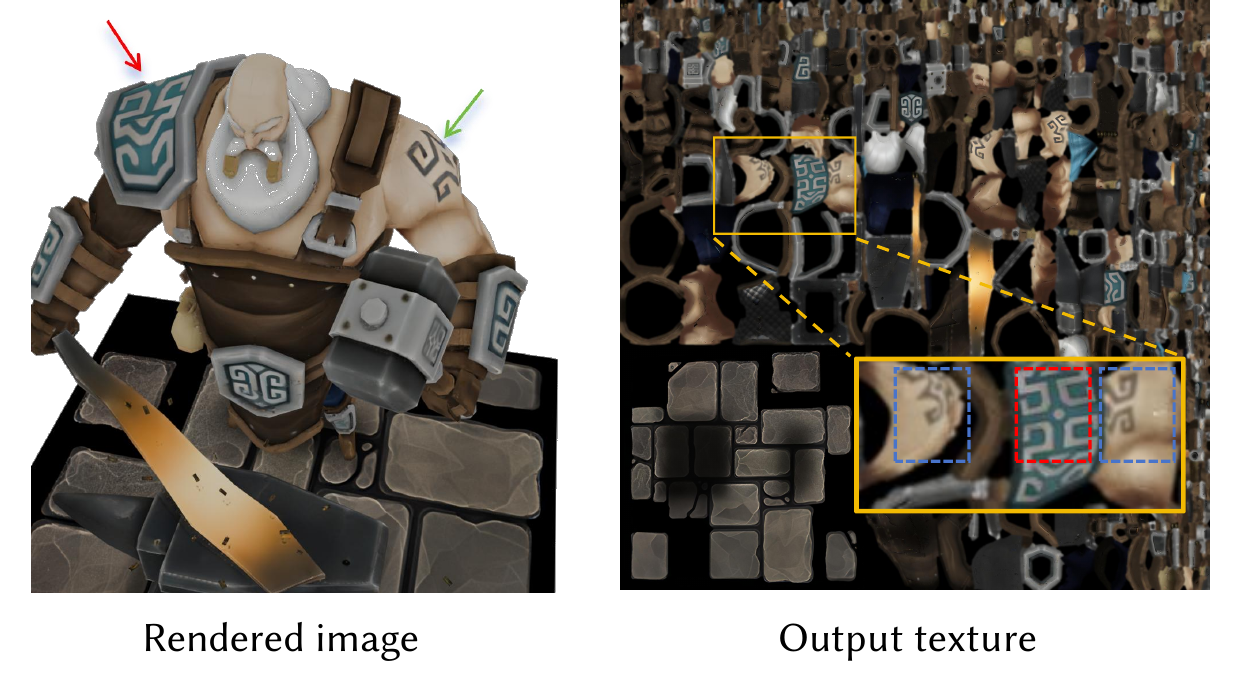}
    \vspace{-1cm}
    \caption{\textbf{Limitation on severely self-occluded geometry.} Our method fails on regions that are largely occluded by other parts of the model. In the rendered image (left), the left shoulder pattern (green arrow) is visible and successfully super‑resolved, while the right shoulder pattern (red arrow) is occluded by the armor. In the output texture (right), the blue box marks the corresponding texture region that remains blurry due to the lack of valid observations; the red box indicates the texture of the occluding geometry.}
    \vspace{-.5cm}
    \label{fig:fail case}
\end{figure}

Our framework performs well across diverse assets, yet it has several limitations. First, it struggles with models that exhibit highly complex or self-occluding geometry. A representative failure case is shown in Figure~\ref{fig:fail case}: on a human character, the pattern on the left shoulder (green arrow) is unobstructed and successfully super‑resolved, whereas the pattern on the right shoulder (red arrow) is occluded by the armor.
Consequently, the corresponding texture region (blue box), whose underlying geometry is occluded by the geometry linked to the red box, cannot obtain meaningful updates and remains blurry.
In addition, our method does not yet support PBR. In future work, we will explore more adaptive view sampling strategies to enhance robustness on geometrically complex assets, and we will also extend the framework to support PBR material SR.

\begin{figure}[h]
    \centering
\includegraphics[width=\linewidth]{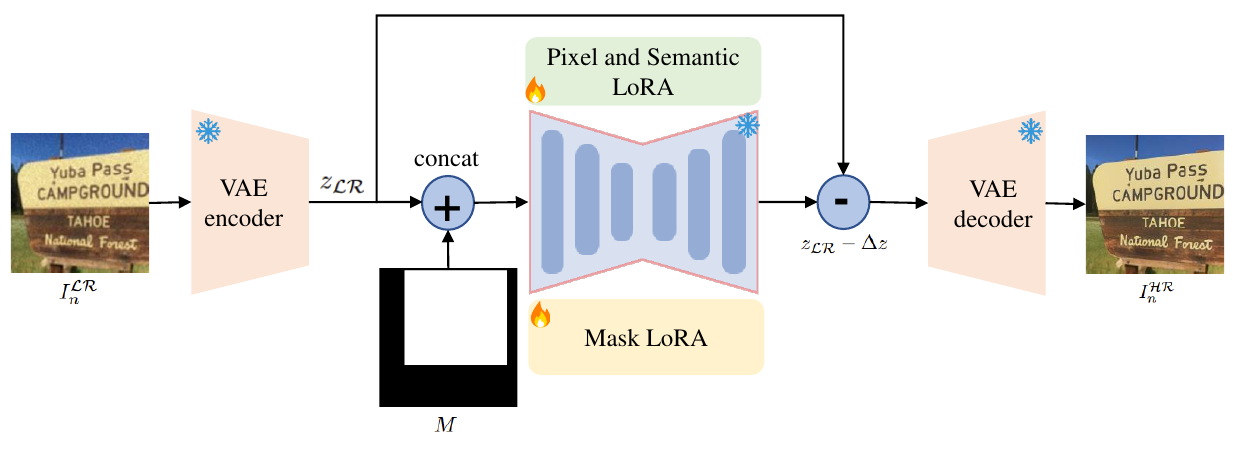}
    \caption{\textbf{Architecture of our diffusion-based local SR module.} The LR image $I_n^{\mathcal{LR}}$ is first encoded into a latent representation $z_{\mathcal{LR}}$ by the frozen VAE encode. This latent is concatenated with the binary mask $M$ and inputed into the U-Net. The U-Net is fine-tuned using two trainable LoRA modules and performs a single-step prediction of a latent residual $\Delta z$, which is subtracted from the original latent $z_{\mathcal{LR}}$. The final, enhanced latent is then decoded by the frozen VAE decoder to produce the HR output $I_n^{\mathcal{HR}}$. Only the LoRA modules and the U-Net's input layer are trained.}
    \label{fig:sr model}
\end{figure}

\section{More implementation details}

\subsection{Datasets}
Due to the lack of sufficient open-source rendered image datasets, we train the module on the LSDIR (Large-Scale Diverse Image Restoration) dataset~\cite{10208419}. LSDIR is well-known for SR task as they contain a vast collection of HR, high-quality images with diverse content. 
During training, random masks are synthetically generated for each image to simulate the local SR task, and then we create pairs of LR and HR image patches using the RealESRGAN degradation module~\cite{9607421}.
For evaluation, we collect a diverse set of high-quality 3D assets from TRELLIS~\cite{xiang2024structured} and Sketchfab. 
LR textures are synthesized by applying Gaussian blur and $4\times$ bicubic downsampling to the original HR textures.

\subsection{Model Training} Our SR module is built upon the public pre-trained Stable Diffusion 2.1-base model~\cite{rombach2022highresolutionimagesynthesislatent} for the $\times4$ SR task. 
The binary mask $M$ is downsampled to match the spatial dimensions of the latent space and applied Gaussian blur to ensure the generated content blends seamlessly with its surroundings.
The Mask LoRA is applied to the shallow encoder blocks of the U-Net and the Pixel-and-Semantic LoRA is applied to the down-sampling blocks, the middle block, and the up-sampling blocks as shown in Figure~\ref{fig:sr model}.
We configure LoRAs with a rank of $r=8$ and a scaling factor of $\alpha=16$. 
We use Adam~\cite{kingma2017adammethodstochasticoptimization} optimizer with a constant learning rate of $5 \times 10^{-5}$ and a batch size of 12. The model is trained for 10,000 steps. The weighting coefficients for our hybrid loss function are empirically set to $\lambda_{mse}=1.0$ and $\lambda_{lpips}=2.0$ to strike a balance between structural fidelity and perceptual realism.

\subsection{The details of the local SR model}
In this section, we provide a more detailed description of the architecture and training procedure for our diffusion-based local SR module and we will release the detailed network source code in the future.
\paragraph{Network Architecture and Data Flow.} Our enhancement module is based on a Stable Diffusion 2.1 model, specifically adapted for single-step, masked SR. The detailed data flow during a single forward pass is illustrated in Figure~\ref{fig:sr model} and proceeds as follows:

\textbf{Step 1. VAE Encoding:} The LR input shading image, $I_n^{\mathcal{LR}}$, is first encoded into a 4-channel latent representation, $z_{\mathcal{LR}}$, using the pre-trained VAE encoder.

\textbf{Step 2. Mask Conditioning:} The binary mask image, $M_n$, is downsampled to match the spatial dimensions of $z_{\mathcal{LR}}$, This downsampled mask is then concatenated with $z_{\mathcal{LR}}$ along the channel dimension, forming a 5-channel tensor that serves as the input to the U-Net. To accommodate this, the U-Net's initial convolutional layer was replaced with a new layer accepting a 5-channel input.

\textbf{Step 3. LoRA-guided Denoising:} The 5-channel tensor is processed by the U-Net, whose generative process is controlled by two distinct, trainable LoRA modules integrated into its attention layers. The Pixel-and-Semantic LoRA module governs the synthesis of high-fidelity content and the restoration of core image structures. The Mask LoRA module enforces spatial conditioning, constraining the synthesis process primarily to the masked regions and promoting a seamless transition with the unmasked context.

\textbf{Step 4. Single-Step Residual Prediction:} The LoRA-guided U-Net performs a single forward pass to predict a latent-space residual, $\Delta z$. This residual is then subtracted from the initial latent representation, $z_{\mathcal{LR}}$, to compute the final high-resolution latent, $z_{\mathcal{HR}}$.

\textbf{Step 5. VAE Decoding:} Finally, the high-resolution latent, $z_{\mathcal{HR}}$, is passed through the pre-trained VAE decoder to produce the final high-resolution output image, $I_n^{\mathcal{HR}}$.
\paragraph{Training Procedure. } The model is trained end-to-end to optimize the weights of the two LoRA modules and the modified layer. The weights of the base Stable Diffusion U-Net and the VAE encoder/decoder are kept frozen. The training objective is a hybrid loss function computed between the output image $I_n^{\mathcal{HR}}$ and the ground-truth HR image $I_n^\mathcal{GT}$. As described in the main paper, this loss is a weighted combination of a masked L2 MSE loss($\mathcal{L}_{mse}$) and a masked LPIPS perceptual loss($\mathcal{L}_{lpips}$). Further details on the training hyperparameters, including learning rate and batch size, are provided in the main paper.

\section{Sensitivity to UV Parameterization}
We further evaluate the robustness of our approach to UV parameterization by testing multiple unwrapping strategies, including Blender's Smart UV Project(45° and 65° angle limits) and xatlas, which produce distinct seam layouts and island shapes. As shown in Table~\ref{tab:uv parameterization}, our method maintains consistent visual quality across all configurations. 
\begin{table}
    \centering
    \begin{tabular*}{\columnwidth}{l @{\extracolsep{\fill}} cccc}
        \toprule
        UV Layout & PSNR $\uparrow$ & SSIM $\uparrow$ & LPIPS $\downarrow$ & DISTS $\downarrow$ \\
        \midrule
        xatlas & 34.69 & 0.9519 & 0.0470 & 0.0911 \\
        45° & 34.70 & 0.9511 & 0.0485 & 0.1037 \\
        60° & 35.00 & 0.9521 & 0.0481 & 0.1003 \\
        \bottomrule
    \end{tabular*}
    \caption{\textbf{Quantitative evaluation of robustness to UV parameterizations.} The minimal variance across all three metrics demonstrates that our method is robust to variations in UV island shapes and artificial seam placements.}
    \label{tab:uv parameterization}
\end{table}
\section{Iterative Texture Refinement}
\begin{figure}
    \centering
    \includegraphics[width=\linewidth]{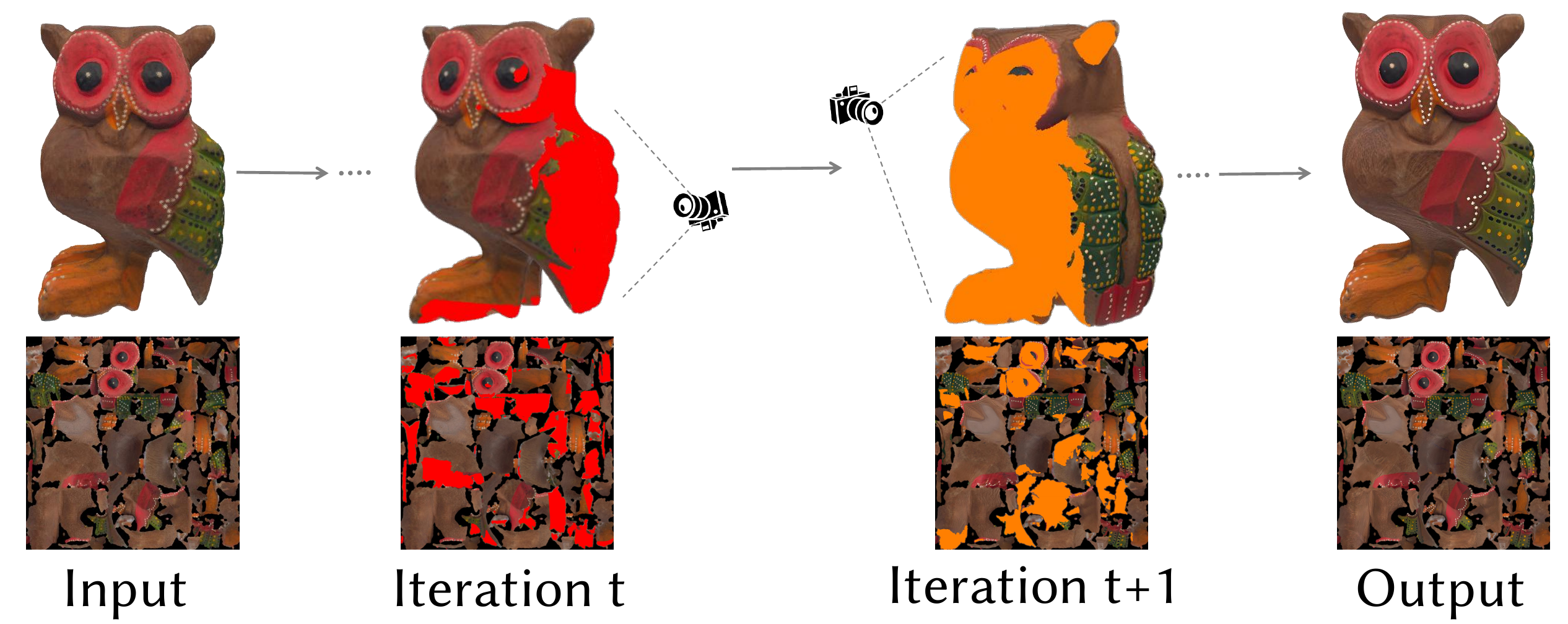}
    \caption{\textbf{Pipeline overview.} From the LR texture, we iteratively SR local regions across different viewpoints (each iteration highlighted in a distinct color) until the final HR texture is produced.}
    \label{fig:iteration}
\end{figure}

Figure~\ref{fig:iteration} illustrates the overview of our iterative texture SR pipeline. Starting with a LR input, our method progressively enhances texture details through a sequence of rendering and SR steps. In each iteration, a specific part of the surface is selected for SR, visualized by the distinct colors on both the 3D mesh and the corresponding UV layout. This process allows the model to focus on local regions from optimal viewpoints, gradually covering the entire surface until a complete HR texture is generated.

\section{The detail of the viewpoint consolidation}
We consolidate island and seam viewpoints using a greedy clustering algorithm governed by two criteria.
The positional threshold $\tau_{\text{pos}}$ is defined as a fixed proportion of the 3D model bounding box diagonal, making it scale‑invariant across different meshes.
A viewpoint is assigned to a cluster only if both its spatial distance to the cluster centroid is below $\tau_{\text{pos}}$ and the cosine similarity between its viewing direction and the cluster's mean direction exceeds $\tau_{\cos}$.
This dual constraint prevents erroneous merges between viewpoints that are spatially proximate but face opposite directions, or those that share similar orientations but are distant.
Then we recompute the visible face set of each merged view through frustum culling, detect and cover uncovered areas with additional camera viewpoints.
\begin{figure*}
    \centering
    \includegraphics[width=\linewidth]{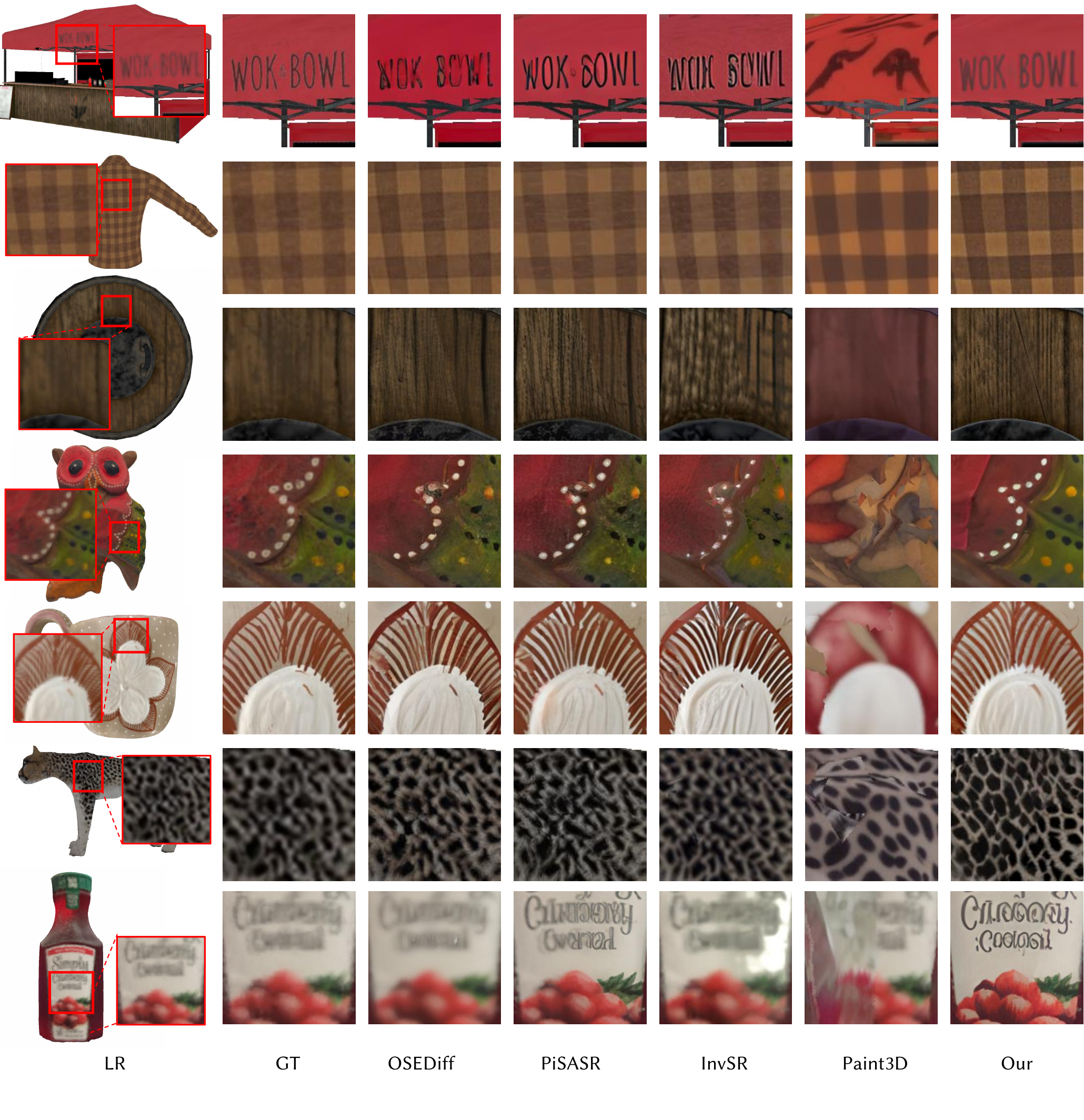}
    \vspace{-1cm}
    \caption{\textbf{Qualitative comparison with additional baseline methods.} We further compare against image SR methods (OSEDiff, PiSASR, InvSR) and a texture generation method (Paint3D). Consistent with the previous evaluation, these baselines also exhibit UV boundary discontinuities or over‑smoothed textures and hallucinated details (Paint3D) that deviate from the GT. In contrast, our method consistently reconstructs sharper, more faithful textures and avoids such artifacts.}
    \label{fig:another baseline}
\end{figure*}
\begin{figure*}
    \centering
    \includegraphics[width=\linewidth]{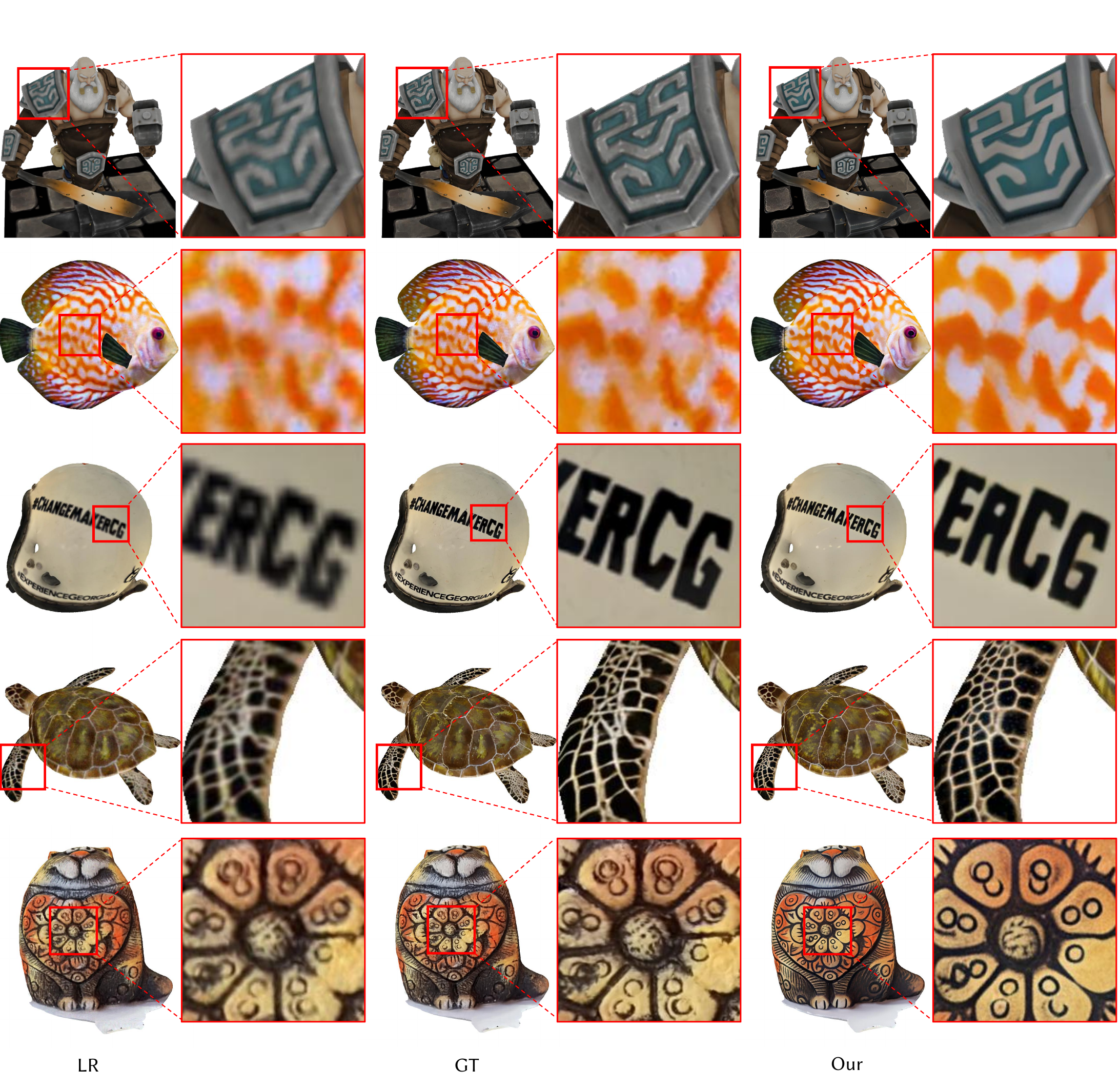}
    \vspace{-0.8cm}
    \caption{\textbf{Qualitative results (3/3)}. Each triplet shows the rendered LR input, our HR output, and the GT. These examples span both structured patterns (text, symbols) and organic surface details (such as turtle and fish skin textures), demonstrating the versatility of our method.}
    \label{fig:more results3}
\end{figure*}
\begin{figure*}
    \centering
    \includegraphics[width=\linewidth]{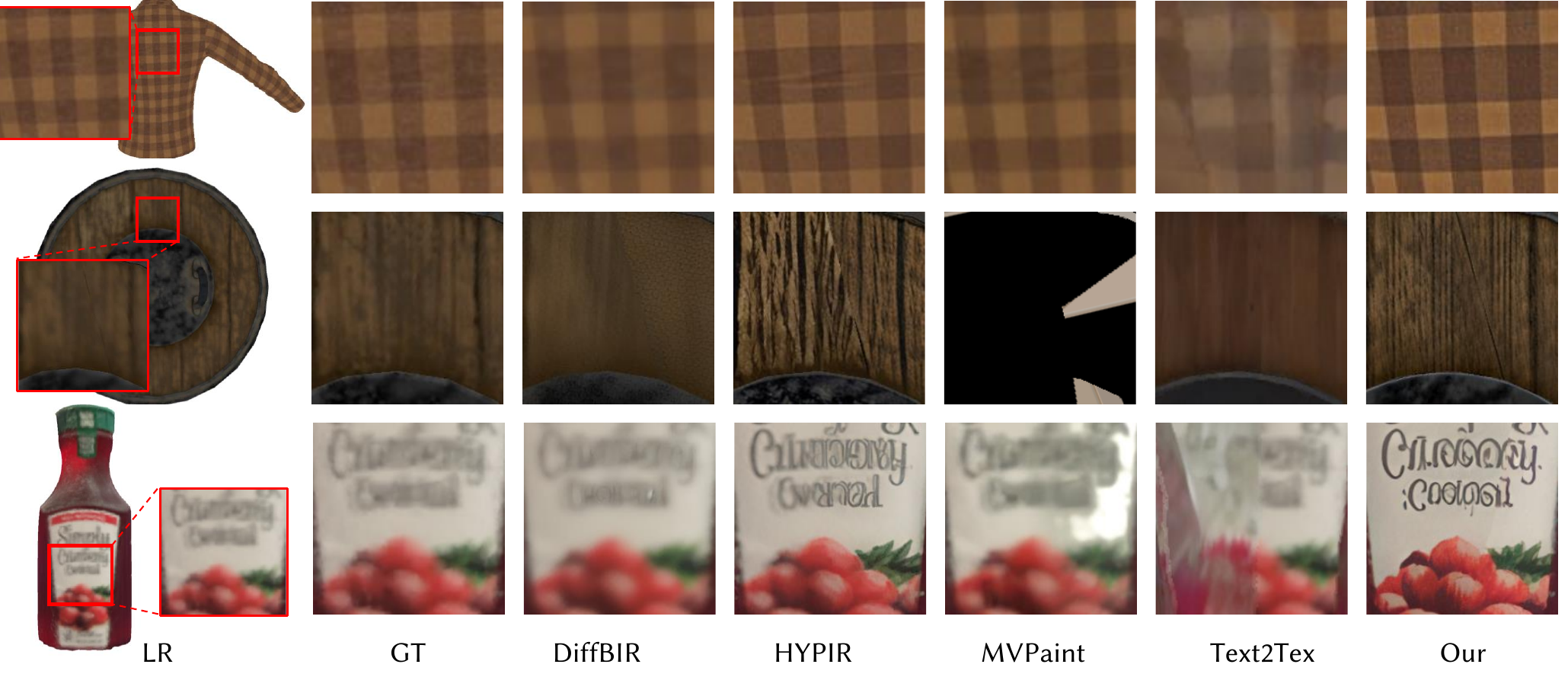}
    \vspace{-0.8cm}
    \caption{Qualitative comparison with SOTA image SR (DiffBIR, HYPIR) and texture generation methods (MVPaint, Text2Tex).}
    \label{fig:baseline2}
\end{figure*}

\end{document}